\documentclass[runningheads]{llncs}

\usepackage{eccv}

\usepackage[table]{xcolor}
\usepackage{xcolor}
\definecolor{mycomment}{RGB}{255,0,0}  
\definecolor{myblue}{RGB}{0,0,255}   
\usepackage{multirow}
\usepackage{eccvabbrv}

\usepackage{graphicx}
\usepackage{subcaption}
\usepackage{booktabs}
\usepackage{color}
\usepackage[accsupp]{axessibility}  

\usepackage{hyperref}

\usepackage{orcidlink}

\begin{document}

\title{Grounding Sim-to-Real Generalization in Robotic Manipulation: An Empirical Study with Vision-Language-Action Models}

\titlerunning{Grounding Sim-to-Real Generalization in VLAs}

\author{Ruixing Jin\inst{1} \and
Zicheng Zhu\inst{1} \and
Ruixiang Ouyang\inst{1} \and
Sheng Xu\inst{1} \and
Bo Yue\inst{1} \and
Zhizheng Wu\inst{1} \and
Guiliang Liu\thanks{Corresponding Author.} \inst{1,2}
}

\authorrunning{Jin et al.}

\institute{
  ${}^{1}$School of Data Science, The Chinese University of Hong Kong, Shenzhen\\
  ${}^{2}$Shenzhen Loop Area Institute
  \\ 
  \email{\{ruixingjin, zichengzhu, ruixiangouyang, shengxu1, boyue\}@link.cuhk.edu.cn} \\
  \email{\{wuzhizheng, liuguiliang\}@cuhk.edu.cn}
}
\maketitle

\begin{abstract}
Learning a generalist control policy for robotic manipulation typically relies on large-scale datasets. Given the high cost of real-world data collection, a practical alternative is to generate synthetic data through simulation. However, the resulting synthetic data often exhibits a significant gap from real-world distributions. While many prior studies have proposed algorithms to bridge the Sim-to-Real discrepancy, there remains a lack of principled research that grounds these methods in real-world manipulation tasks, particularly their performance on generalist policies such as Vision-Language-Action (VLA) models.
In this study, we empirically examine the primary determinants of Sim-to-Real generalization across four dimensions: multi-level domain randomization, photorealistic rendering, physics-realistic modeling, and reinforcement learning updates.
To support this study, we design a comprehensive evaluation protocol to quantify the real-world performance of manipulation tasks. The protocol accounts for key variations in background, lighting, distractors, object types, and spatial features. Through experiments involving over 10k real-world trials, we derive critical insights into Sim-to-Real transfer. To inform and advance future studies, we release both the robotic platforms and the evaluation protocol for public access to facilitate independent verification, thereby establishing a realistic and standardized benchmark for robotic manipulation policies.
\end{abstract}
%
\section{Introduction}

Developing precise and scalable robotic manipulation policy represents a key milestone toward realizing artificial general intelligence (AGI)~\cite{Smith2012DualArm}. 
To achieve this goal, Vision-Language-Action (VLA) models have emerged as a principled design for building general-purpose robotic agents, enabling an end-to-end mapping from visual observations and language instructions to continuous motor actions~\cite{ma2024survey,zheng2025survey}.
VLA methods typically adapt multi-modal foundation models~\cite{chen2023palixscalingmultilingualvision, Touvron2023Llama2O, Li2025Eagle2B} to embodied control and fine-tune them on diverse robotic demonstrations~\cite{Padalkar2023OpenXR, Khazatsky2024DROIDAL}, thereby exhibiting compositional reasoning capabilities from foundation models~\cite{RT2, openvla, pi0, Nvidia2025GR00TNA,Team2024OctoAO}. This formulation allows a single policy to perform multi-task manipulation across heterogeneous environments, demonstrating promising language grounding and cross-task generalization~\cite{Zhao2023ASO, Xu2026FromRT}.

\begin{figure}[t]
    \centerline{\includegraphics[width=1\linewidth]{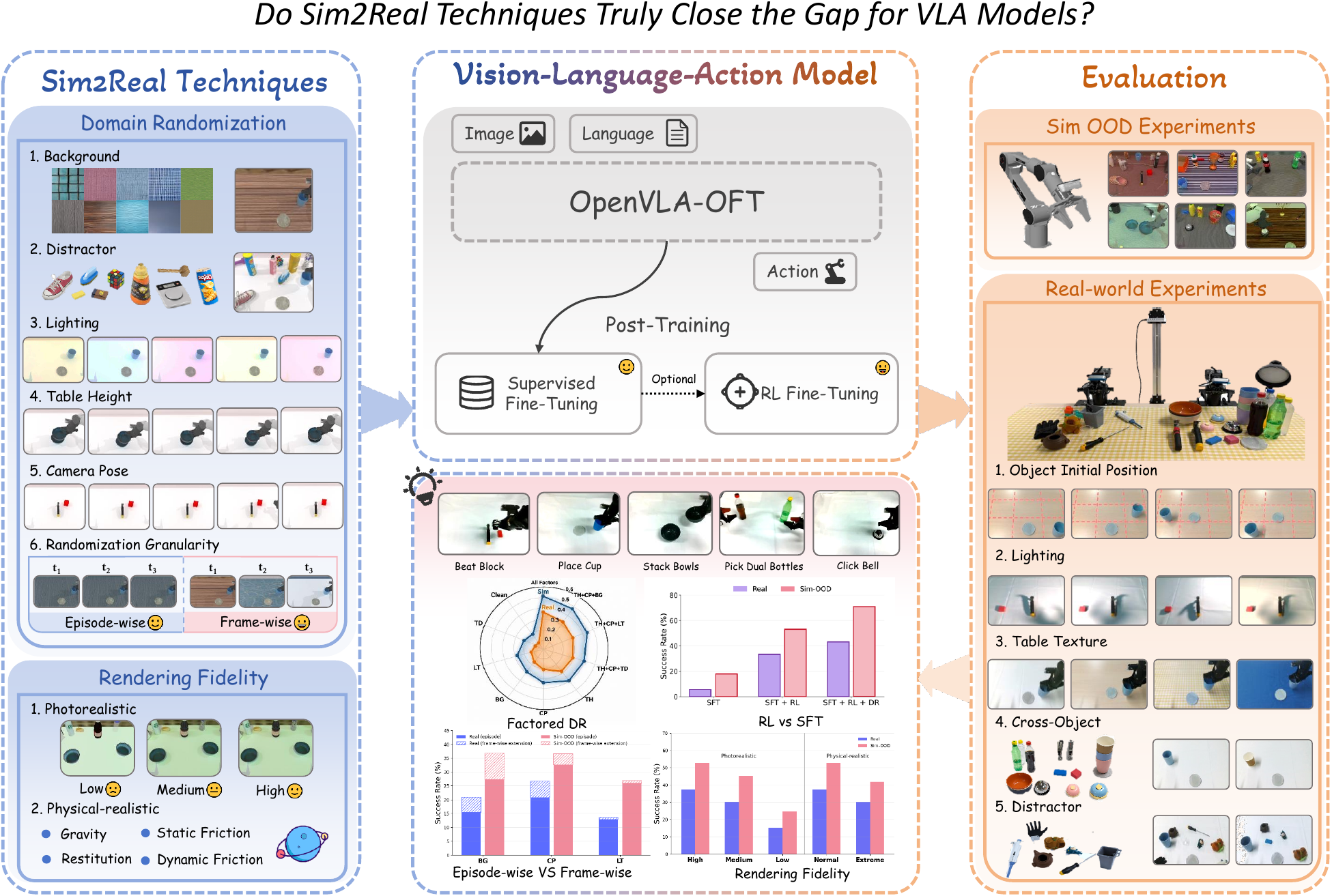}}
    \caption{\small Overview of our framework for analyzing Sim2Real generalization in Vision-Language-Action (VLA) models. 
We study how different Sim2Real techniques, including domain randomization, rendering fidelity, and reinforcement learning fine-tuning, influence generalization across Vision, Semantics, and Execution under both simulation OOD and real-world evaluations.}
    \label{fig:teaser}
\end{figure}

As a learning-based framework for robotic manipulation, VLA training typically follows a data-driven pipeline. However, this process requires a substantial amount of robotic operation trajectories, which are fundamentally scarce, costly to acquire, and inherently hardware-specific. 
These characteristics make it difficult to scale across diverse tasks, robotic platforms, and real-world scenarios~\cite{DBLP:conf/iclr/ChenHJLW22}.
To address these challenges, recent studies~\cite{mandlekar2023mimicgen,deng2025graspvla,Liu2025DexScale,zhao2026simreal,Lin2026DyGROVLACS} have explored training VLA models using large-scale simulated data generated by diverse simulation platforms. These platforms include rule-based simulators, such as MuJoCo~\cite{todorov2012mujoco} and Isaac Lab~\cite{mittal2025isaaclab}, as well as learning-based environments built upon world models~\cite{team2025gigaworld}. 
Compared to data collection in real-world environments, synthetic data can be produced efficiently at scale. However, significant discrepancies in dynamics, kinematics, sensor characteristics, or other factors between simulated and real settings result in a pronounced simulation-to-reality (Sim2Real) gap. As a result, VLA models trained in simulation often exhibit degraded performance when deployed in real-world scenarios~\cite{Robotwin2.0,Din2025VisionLA,Nasiriany2024RoboCasaLS}.

The ability to deploy policies trained in simulation to the physical world is therefore essential for scalable and practical robotics. 
Existing Sim2Real approaches primarily fall into four categories:
1) {Domain randomization}~\cite{Chen2022UnderstandDR} improves robustness to out-of-distribution (OOD) real-world conditions by introducing diverse random perturbations during simulation training.
2) {Domain adaptation}~\cite{Bousmalis2018DA,josifovski2025safe,Zhang2019Goggles} reduces the discrepancy between simulated and real domains by aligning them within a shared feature space.
3) {Photo- or physics-realistic rendering} mitigates the Sim2Real gap by increasing the visual fidelity of observations and the physical accuracy of environmental dynamics.
4) {Reinforcement fine-tuning (RFT)}~\cite{simplevla} further improves VLA models by optimizing them through interaction with simulated environments using reinforcement learning (RL).
However, these approaches are often evaluated independently on general robotics tasks or control models. It remains unclear \emph{which specific factors influence, or to what extent they contribute to, the Sim2Real transferability of VLA models in robotic manipulation tasks}.
Such a lack of mechanistic understanding limits the principled optimization of Sim2Real strategies and hinders systematic diagnosis of VLA model failures under real-world deployment.

In this work, we address this question through a factorized empirical analysis of the zero-shot Sim2Real performance of VLA models, where models are trained entirely in simulation and directly deployed to real-world tasks without any real-world data fine-tuning~\cite{zhao2026simreal}. 
To evaluate \textbf{zero-shot Sim2Real transfer}, we build a unified benchmark for VLA systems based on the RoboTwin 2.0 simulation framework~\cite{Robotwin2.0}. 
The benchmark includes a set of representative dual-arm manipulation tasks covering both short-horizon precision interactions and multi-stage behaviors. 
In addition to simulation OOD evaluation, we conduct real-world tests under controlled variations, including background textures, lighting perturbations, object instances, distractors, and spatial position changes across predefined grid layouts. 

Within this unified framework, we conduct a factorized comparison of Sim2Real methods on zero-shot transfer performance, examining the effects of domain randomization factors, randomization granularity, photorealistic rendering fidelity, and RL fine-tuning. Through a massive real-world evaluation including more than 10,000 real-world trials, our study offers the following insights:
\begin{itemize}

    \item \textbf{Spatial features speak louder than appearance.} Augmenting spatial features, such as table height and camera pose, yields larger improvements than purely visual perturbations like background textures or lighting. This suggests that spatial variation enhances the policy’s ability to adapt to different spatial configurations and strengthens the coupling between visual perception and motor execution.

    \item \textbf{Granularity matters.} 
    Frame-wise domain randomization yields better zero-shot transfer than episode-wise strategies by improving the model's attention to task-relevant objects rather than background variations.

    \item \textbf{Simulation fidelity enhances Sim2Real transfer.} Our results reinforce the conclusion that higher levels of photorealistic and physics-based fidelity consistently improve Sim2Real performance; however, the magnitude of improvement diminishes as the simulation fidelity reaches a threshold.
    \item \textbf{RL enhances robustness to distribution shifts.} Compared to SFT-only policies, RFT demonstrates greater resilience to object variations and environmental perturbations. When combined with structured domain randomization, the performance gains are further amplified, highlighting the complementary benefits of policy optimization and environmental diversity.
\end{itemize}

Inspired by RoboChallenge~\cite{yakefu2025robochallenge}, we develop an online system that offers open access to standardized protocols, experimental settings, and real-world deployment interfaces of bimanual robots. This platform ensures result reproducibility and enables practitioners to gain deeper insights into Sim2Real transfer. It supports continued investigation and systematic benchmarking of VLA models in physical environments.

\section{Related Work}
\subsection{Vision-Language-Action (VLA) Models}
Recently, with the advancement of large language models (LLMs) and vision-language models (VLMs)~\cite{chen2023palixscalingmultilingualvision, Touvron2023Llama2O, Li2025Eagle2B}, training Vision-Language-Action (VLA) models on large-scale robotic datasets~\cite{Padalkar2023OpenXR, Khazatsky2024DROIDAL} has become a significant research direction in the robotics field. Early works such as RT-1~\cite{RT1} and RT-2~\cite{RT2} demonstrated the feasibility of training transformer architectures on large-scale robotic datasets by adding action heads to VLMs. Similarly, open-source models such as OpenVLA~\cite{openvla}, and CogACT~\cite{CogACT} have also proven that VLA architectures can achieve competitive performance across diverse hardware platforms. Meanwhile, recent studies have continuously optimized model architectures; for instance, $\pi_0$~\cite{pi0} adopted flow matching, and OpenVLA-OFT~\cite{openvla-oft} altered the decoding method for action tokens. However, these works primarily focus on simulated environments or deployment fine-tuned on real-world data. Our research will focus on empirical studies of Sim2Real transfer for VLA models.
\subsection{Techniques for Sim-to-Real Transfer in Robotics}
To bridge the gap between simulation and real-world deployment, several mainstream Sim2Real techniques have been widely studied in robotics. Representative approaches include domain randomization, domain adaptation, improving rendering realism, and RL-based post-training.

\noindent\textbf{Domain Randomization (DR)} remains the foundational approach for zero-shot transfer~\cite{zhu2025domainrandomizationobjectdetection,8202133, 11042418}, which expands the simulated distribution by randomizing visual and physical parameters during training to encompass real-world variations. Tobin et al.~\cite{8202133} proposed that randomizing textures, lighting, and camera parameters enables zero-shot transfer for visual grasping tasks by deep neural networks. Subsequently, DR has been widely adopted in object detection~\cite{11042418}, robotics~\cite{sim2real}, autonomous driving~\cite{9025396,10.1145/3487075.3487177}, and other fields. However, DR is typically applied monolithically~\cite{zhao2026simreal, Robotwin2.0, zhu2025domainrandomizationobjectdetection}, where multiple factors are randomized simultaneously, making it difficult to unravel their individual contributions. Recent studies also have determined relevant parameters by automatic learning~\cite{OpenAI2019ADR}, active exploration~\cite{Mehta2019ADR}, Bayesian update~\cite{Muratore2021NPDR,Muratore2021DR}, offline inference~\cite{Tiboni2023DROPO} and continual learning~\cite{Josifovski2024CDR}. In this work, we address this gap through a factorized empirical study that disentangles and ranks key domain randomization factors to identify their impact on zero-shot Sim2Real transfer for VLA models. 


\noindent\textbf{Photo/Physics Realistic Rendering} provides a complementary strategy to reduce visual Sim-to-Real discrepancies by improving the realism of simulated observations. Instead of expanding the training distribution as in domain randomization, photorealistic simulation aims to narrow the visual reality gap at its source by generating high-fidelity synthetic data. Recent advances in neural scene reconstruction and rendering further enable realistic simulation environments derived from real-world captures~\cite{3DGaussian,4DGussian}. In robotics, modern simulation platforms support high-fidelity GPU-accelerated rendering and physics simulation~\cite{lsaac,spain}, making photorealism increasingly practical for large-scale training. In this work, we treat rendering fidelity as a controlled experimental factor and systematically evaluate its effect on zero-shot Sim2Real transfer.

\noindent\textbf{Reinforcement Fine-Tuning (RFT)} has emerged as a powerful paradigm for enhancing the capabilities of foundation models~\cite{Liu2025ReinforcementLM}. In large language models (LLMs) and vision-language models (VLMs), RL has been shown to significantly improve reasoning abilities~\cite{shao2024deepseekmathpushinglimitsmathematical} and align outputs with human preferences~\cite{DBLP:journals/corr/abs-2312-14925}. Beyond standard supervised fine-tuning (SFT), RL fine-tuning can unlock out-of-distribution generalization and stronger reasoning performance~\cite{whatcanRLbringtovla, SFTMemoriesRLgen}. These benefits have recently extended to vision-language-action (VLA) models, where RL fine-tuning has been shown to improve policy performance in both simulation and real-world settings~\cite{whatcanRLbringtovla, simplevla}. In this work, we further examine how RL fine-tuning influences zero-shot Sim2Real generalization in VLA models.
\subsection{Real-Robot Evaluation}
While simulation platforms are gradually maturing, real-robot evaluation remains indispensable. RoboChallenge \cite{RoboChallenge} identifies this as a significant challenge in robotics and has constructed a large-scale online real-robot evaluation infrastructure, introducing multiple tabletop manipulation tasks. In establishing real-world test sets, Xie et al.~\cite{DecomposingTG} discovered that different environmental factors have varying impacts on model performance in the real world, and these factors are largely independent of each other, which means most pairs of factors exhibit no compounding effects. This finding also serves as an important basis for our design of real-world test sets. In this work, we design a controlled benchmark to comprehensively evaluate the zero-shot Sim-to-Real capability of VLA models.
\section{Preliminaries}

\subsection{VLA Models for Robotic Manipulation}

Robotic manipulation in the Sim2Real setting can be formulated as a Partially Observable Markov Decision Process (POMDP) 
$\mathcal{M} = (\mathcal{S}, \mathcal{A}, \mathcal{O}, P_T, R, \mu_0, \gamma)$ where: 1) The state $s_t \in \mathcal{S}$ captures the semantic information of a scene, encompassing the configuration (e.g., layouts, appearance, and physical characteristics) of various types of objects and the robots.
2) The action $a_t \in \mathcal{A}$ corresponds to low-level robot control commands. 
In our work, actions are parameterized as continuous target joint angles. 
3) The observation $o_t \in \mathcal{O}$ represents the perceptual signals captured by sensors. These observations are typically non-Markovian and provide only partial information about the current state.
4) The transition function $P_T$ describes the physical dynamics that map the current state $s_t$ and executed action $a_t$ to the next state $s_{t+1}$. 
5) The reward function $R(s,a)$ measures task performance and, in many robotic manipulation settings, is defined by task goals (e.g., language instructions) and often simplified as a binary success signal indicating task completion~\cite{simplevla, yue2025real}.
6) $\mu_0$ denotes the initial state distribution, and $\gamma \in (0,1]$ is the discount factor that balances immediate and future rewards.

The VLA policy is conditioned on a language instruction $l$ that specifies the task goal and learns a mapping 
$\pi(a_{t:t+M} \mid o_t, l)$ to predict a short-horizon sequence of future actions from the current observation and instruction~\cite{openvla-oft}. 
The objective is to maximize expected task success under instruction-conditioned trajectories, allowing a unified policy to map high-level semantic commands to low-level motor control. 
This language interface enables generalization across diverse manipulation goals, object variations, and spatial configurations~\cite{openvla, openvla-oft, pi0}.

\subsection{Zero-Shot Sim2Real Paradigm for VLA Models}
Let $\hat{\mathcal{M}} = (\mathcal{S}, \mathcal{A}, \mathcal{O}, P, R, \mu_0, \gamma)$ denote the real-world environment, 
and let $\mathcal{M}$ denote the corresponding simulated environment. 
In simulation, we collect trajectories 
$\tau = (o_0, a_0, \dots, o_T, a_T, l)$ 
generated by interacting with $\mathcal{M}$, where $l$ specifies the manipulation task.

The policy is optimized in simulation as:
\begin{equation}
\pi^* =
\arg\max_{\pi}
\mathbb{E}_{\tau \sim \mathcal{M}}
\big[
J(\tau, \pi; l)
\big].
\end{equation}

We present a unified framework for analyzing zero-shot Sim2Real transfer in VLA systems. Our method organizes the problem into two components: policy learning and simulation design. 
\subsubsection{Supervised Fine-Tuning (SFT)}
We first train the VLA policy in simulation using supervised behavioral cloning. 
Given a demonstration dataset 
$\mathcal{D} = \{ (o_t^{(i)}, a_{t:t+M}^{(i)}, l^{(i)}) \}$, 
where $i$ denotes the data index, $o_t$ denotes the current observation, and $l$ the language instruction, the policy predicts a short-horizon sequence of future continuous actions conditioned on multimodal inputs.

Actions are parameterized as continuous low-level control commands 
(e.g., joint targets and gripper states), 
forming an action vector $a_{t:t+M} \in \mathbb{R}^{(M+1)d_a}$ over a prediction horizon of $M+1$ steps. 
The policy directly outputs continuous action values rather than discretized tokens.

The supervised objective minimizes the L1 regression loss~\cite{openvla-oft}:
\begin{equation}
\mathcal{L}_{\text{SFT}}(\theta)
=
\mathbb{E}_{(o_t, a_{t:t+M}, l) \sim \mathcal{D}}
\left[
\left\|
\pi_\theta(o_t, l) - a_{t:t+M}
\right\|_1
\right],
\end{equation}
where $\theta$ denotes the learnable model parameters.

\subsubsection{Reinforcement Learning Fine-Tuning}
To further enhance robustness, we optionally perform reinforcement learning (RL) fine-tuning in simulation. In our study, we adopt Group Relative Policy Optimization (GRPO)~\cite{shao2024deepseekmathpushinglimitsmathematical} as used in SimpleVLA-RL~\cite{simplevla}. GRPO optimizes the policy by comparing relative returns within a group of sampled trajectories, encouraging actions that outperform group-level baselines while stabilizing training for large VLA models.

Given a task-conditioned reward function ${R}(s_t, l)$, we optimize the policy using a group-relative objective. For each update, we sample $G$ trajectories $\{\tau_i\}_{i=1}^{G}$ from the current policy and compute trajectory-level returns $R_i$. Instead of using a fixed reference model with KL regularization, we follow the DAPO-style modification and remove the KL penalty to encourage exploration~\cite{Yu2025DAPOAO}.

The policy is optimized via the following clipped objective:
\begin{equation}
\mathcal{J}(\theta)
=
\mathbb{E}_{\{a_t\}\sim\pi_{\theta_{\text{old}}}}
\left[
\frac{1}{G}
\sum_{i=1}^{G}
\frac{1}{|\tau_i|}
\sum_{t}
\min
\left(
r_{i,t}(\theta)\hat{A}_i,
\,
\text{clip}\big(r_{i,t}(\theta),1-\epsilon,1+\epsilon\big)\hat{A}_i
\right)
\right],
\end{equation}
where
\begin{equation}
r_{i,t}(\theta)
=
\frac{\pi_\theta(a_{i,t} \mid o_{i,t})}
{\pi_{\theta_{\text{old}}}(a_{i,t} \mid o_{i,t})},
\quad
\hat{A}_i
=
\frac{R_i - \text{mean}(\{R_j\}_{j=1}^{G})}
{\text{std}(\{R_j\}_{j=1}^{G})}.
\end{equation}
Here $r_{i,t}(\theta)$ denotes the policy likelihood ratio, $\hat{A}_i$ is the normalized advantage computed from trajectory returns, and $\epsilon$ is a clipping hyperparameter that constrains $r_{i,t}(\theta)$ within $[1-\epsilon,\,1+\epsilon]$ to stabilize policy updates. Only trajectory groups containing both successful and unsuccessful rollouts are used for updates, ensuring meaningful relative comparisons within each group~\cite{simplevla}.

Notably, Sim2Real transfer in our setting is conducted in a zero-shot manner, where no real-world demonstrations are used during training and the learned policy must directly generalize to the real environment $\hat{\mathcal{M}}$. However, discrepancies between $\mathcal{M}$ and $\hat{\mathcal{M}}$ introduce distribution shifts in visual appearance and geometry. To mitigate this gap, we focus on simulation-side techniques that improve robustness without requiring real-world adaptation data. While domain adaptation methods align simulated and real observations using real-world data, they fall outside the zero-shot setting considered in this work. Instead, we adopt two complementary strategies for constructing simulation training data: \textbf{Domain Randomization}, which exposes the policy to diverse environment configurations, and \textbf{Rendering Fidelity}, which improves the realism of simulated observations, including photorealism and physical realism.

\subsubsection{Domain Randomization (DR).}
We train policies over a distribution of simulator parameters rather than a fixed environment~\cite{DBLP:conf/iclr/ChenHJLW22}. 
Let $x$ denote an observation generated by the simulator, and let $\xi$ represent the simulator parameters controlling the scene configuration. 
In a fixed simulator, observations follow $x \sim p_\xi(x)$. 

Under DR, simulator parameters are sampled from a distribution $\xi \sim p(\xi)$, leading to a training distribution $x \sim \int p(x \mid \xi) p(\xi) d\xi$. 

The learning objective for a policy $f_\phi$ becomes 
\begin{equation}
\mathcal{L} = \mathbb{E}_{\xi \sim p(\xi)} \left[ \mathbb{E}_{x \sim p(x|\xi)} \ell(f_\phi(x), y) \right]. 
\end{equation} 
The objective of DR is not to exactly match the real distribution $p_{\text{real}}(x)$, but to expand the synthetic distribution such that $\mathrm{supp}(p_{\text{real}}) \subseteq \mathrm{supp}(p_{\text{DR}})$, so that real-world observations appear as a special case within the randomized synthetic distribution.

Following structured domain randomization principles from prior manufacturing DR literature~\cite{zhu2025domainrandomizationobjectdetection, 11042418}, we factorize domain randomization into structured components as $\xi = \{ \xi_{\text{bg}}, \xi_{\text{dist}}, \xi_{\text{cam}}, \xi_{\text{light}}, \xi_{\text{table}} \}$.

\paragraph{Domain Randomization Factors.}
We decompose domain randomization into five structured components to analyze their individual contributions: \textbf{Background} ($\xi_{\text{BG}}$), which randomizes wall and table textures from a predefined image set; \textbf{Table Distractor} ($\xi_{\text{TD}}$), which samples the number, mesh, pose, and texture of irrelevant objects while avoiding overlap with task-relevant ones; \textbf{Camera Pose} ($\xi_{\text{CP}}$), which applies random translational offsets to the head camera position at initialization; \textbf{Lighting} ($\xi_{\text{LT}}$), which randomizes light color and position within scene bounds; and \textbf{Table Height} ($\xi_{\text{TH}}$), which samples table height from a predefined range.

\paragraph{Randomization Granularity.}
We further study the temporal scope of randomization by comparing two strategies: \textbf{episode-wise}, where factors are sampled once per episode and fixed during rollout, and \textbf{frame-wise}, where factors are resampled at every simulation step.
\subsubsection{Rendering Fidelity.}
Beyond domain randomization, the fidelity of both visual rendering and physics simulation plays a critical role in bridging the Sim2Real gap. \textbf{Photorealism} influences the statistical properties of simulated observations by improving light transport, shadow consistency, and global illumination, producing images that more closely resemble real-world sensor data~\cite{3DGaussian}. In practice, we vary rendering configurations—including ray tracing (RT), samples per pixel, path depth, and denoising—to control the level of visual realism. \textbf{Physical realism} affects the accuracy of object dynamics and contact interactions during manipulation. We vary key physics parameters such as gravitational acceleration, static/dynamic friction coefficients, and restitution, which govern object motion and contact behavior in simulation. Together, these factors determine how closely the simulated environment approximates real-world conditions.
\subsection{Problem Formulation: Empirical Study for Sim2Real VLAs}

Real-world robotic data is scarce, expensive to collect, and highly environment-specific~\cite{DBLP:conf/iclr/ChenHJLW22}, making large-scale training of Vision-Language-Action (VLA) models using real demonstrations impractical. As a result, simulation becomes the primary scalable source of training data.

However, discrepancies between simulated and real environments often cause policies trained in simulation to degrade when deployed in the physical world~\cite{Robotwin2.0,Din2025VisionLA,Nasiriany2024RoboCasaLS}. Although some sim-to-real techniques, such as domain randomization, have shown promising results in robotics~\cite{Mehta2019ActiveDR,Muratore2020DataEfficientDR,zhao2026simreal}, its role in large-scale VLA models remains insufficiently understood. Compared with conventional control policies, VLA models jointly integrate visual perception, language grounding, and long-horizon decision making, which may introduce new sensitivities to simulation–reality discrepancies. 

This motivates a systematic empirical study to disentangle how individual simulation design factors affect Sim2Real transfer in VLA models.

\section{Experiments}
\subsection{Evaluation Protocol.}
We conduct a systematic study of zero-shot simulation-to-real generalization under controlled distribution shifts. 
All experiments use the OpenVLA-OFT policy~\cite{openvla-oft} trained with supervised fine-tuning (SFT) in simulation. 
Within this framework, we evaluate how structured domain randomization, randomization granularity, and rendering fidelity affect Sim2Real transfer in both simulation and real-world environments.
\subsubsection{Simulation Dataset.}
All policies are trained in RobotWin 2.0~\cite{Robotwin2.0}, a physics-based manipulation simulator that supports configurable lighting, camera pose, table height, background textures, and distractor objects.

We evaluate five manipulation tasks with varying horizon lengths and spatial complexity. For each task and each training setting, we collect 100 demonstration trajectories in simulation. 

The simulated robot embodiment uses the Cobot Magic platform to ensure consistency with real-world deployment~\cite{Robotwin2.0}. Visual input is provided by a single forward-facing Intel RealSense D435 RGB camera at 640×480 resolution, adopted for simplicity and clearer interpretation of experimental results. Camera intrinsics remain fixed across experiments, while camera pose may be randomized depending on the experimental condition.
\subsubsection{Policy Architecture} 
We base our study on OpenVLA-OFT~\cite{openvla-oft}, a state-of-the-art Vision-Language-Action model that enhances the original OpenVLA~\cite{openvla} with optimized designs for efficient and robust control. Built on a fused visual encoder and a Llama2 7B language backbone~\cite{Touvron2023Llama2O}, OpenVLA-OFT incorporates two key improvements: (1) Parallel decoding with action chunking, which replaces autoregressive generation to enable single-pass prediction of multiple future actions. (2) Continuous action representation with L1 regression, which avoids information loss from discretization and enables finer-grained control.

\subsubsection{Simulation OOD Evaluation.}
To measure robustness within simulation, we construct Out-of-Distribution (Sim-OOD) environments by introducing unseen variations along individual factors. These include unseen background and table textures, unseen distractor objects, table height perturbations, camera pose shifts, and lighting disturbances.

Each factor is varied independently to isolate its effect on performance. Sim OOD evaluation serves as a controlled proxy for real-world generalization.
\subsubsection{Real-World Evaluation Setup.}

\begin{figure}[t]
    \centerline{\includegraphics[width=1\linewidth]{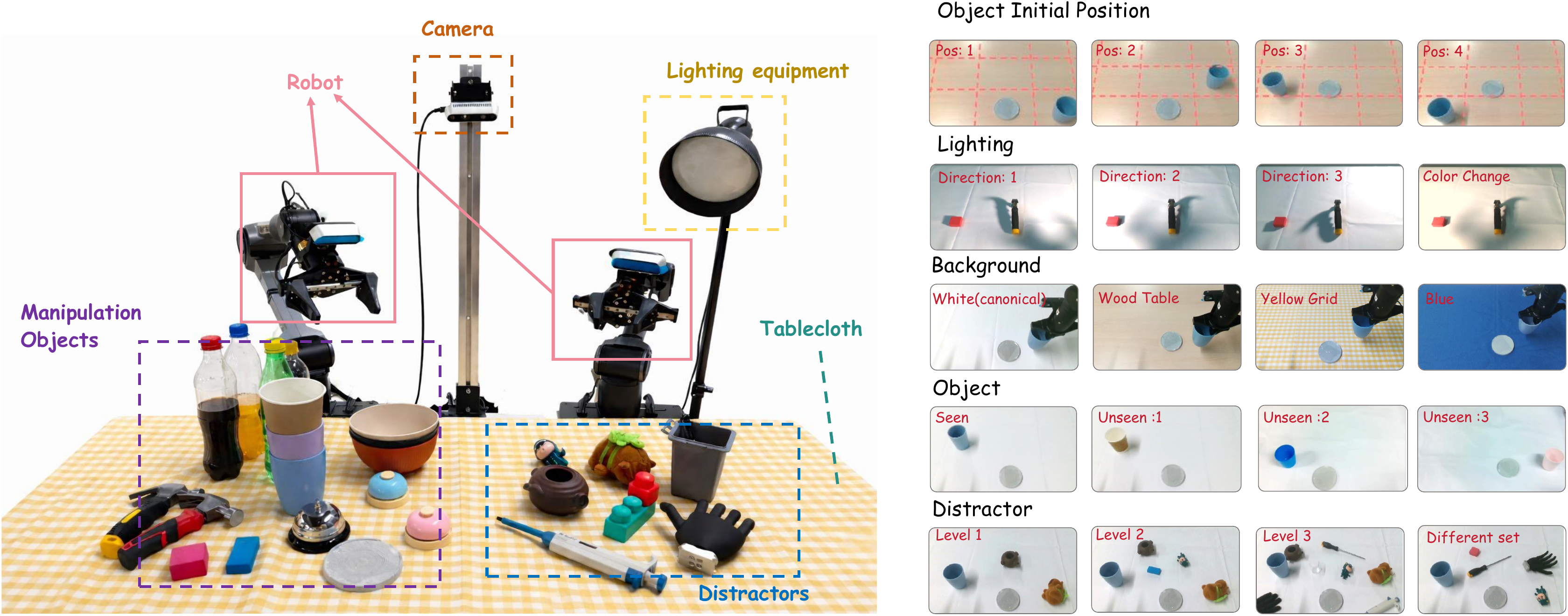}}
   \caption{\small Real-world manipulation setup and randomized factors used in our experiments. Left: the physical platform with robot arms, camera, lighting, objects, and distractors. Right: examples of variations including object positions, lighting, backgrounds, object instances, and distractor configurations.}
    \label{fig:real-world}
\end{figure}

Zero-shot real-world evaluation is conducted on a physical Piper robot platform equipped with a single RealSense D435 RGB camera (640×480 input resolution). An overview of the real-world setup is shown in~\cref{fig:real-world}. The workspace consists of an adjustable-height table with replaceable tablecloth textures and configurable lighting. 

We evaluate robustness across multiple real-world factors:

\begin{itemize}
\item \textbf{Background variation}: three table textures (wood, blue fabric, yellow grid).
\item \textbf{Lighting variation}: three distinct light positions with varying illumination colors.
\item \textbf{Object variation}: both \textit{seen} and \textit{unseen} object instances within the same task category, evaluating generalization to novel appearances.
\item \textbf{Distractor presence}: task-irrelevant objects placed in the scene at three difficulty levels (2, 4, and 8 distractors), sampled from a separate object set distinct from task objects.
\item \textbf{Spatial generalization}: evaluation on unseen positions within a predefined $3 \times 3$ grid.
\end{itemize}

For each task, we first evaluate the policy under a canonical base configuration. We then systematically vary one factor at a time while keeping all other factors fixed. For each configuration, we evaluate the task under multiple object positions by applying small random shifts, and report the resulting success rate across these variations.

\subsection{Domain Randomization for Sim2Real Generalization}
\begin{table}[t]
\caption{Factorized domain randomization results under Simulation OOD and zero-shot real-world evaluation across five tasks.}
\label{tab:factorized_results}
\centering
\resizebox{\columnwidth}{!}{
\begin{tabular}{lcc|cc|cc|cc|cc}
\toprule
 & \multicolumn{2}{c|}{Click Bell}
 & \multicolumn{2}{c|}{Place Empty Cup}
 & \multicolumn{2}{c|}{Beat Block Hammer}
 & \multicolumn{2}{c|}{Stack Bowls Two}
 & \multicolumn{2}{c}{Place Dual Bottles} \\
Data setting
 & Sim-OOD & Real
 & Sim-OOD & Real
 & Sim-OOD & Real
 & Sim-OOD & Real
 & Sim-OOD & Real \\
\midrule
Clean
 & 14\% & 2.7\%
 & 11\% & 5.4\%
 & 17\% & 0.0\%
 & 36\% & 26.2\%
 & 19\% & 1.5\% \\

BG
 & 23\% & 11.5\%
 & 15\% & 10.2\%
 & 30\% & 2.7\%
 & 42\% & 40.4\%
 & 27\% & 12.3\% \\

LT
 & 22\% & 12.3\%
 & 12\% & 8.6\%
 & 29\% & 4.6\%
 & 43\% & 31.5\%
 & 24\% & 7.3\% \\

TD
 & 15\% & 3.1\%
 & 12\% & 6.9\%
 & 23\% & 0.0\%
 & 38\% & 27.3\%
 & 21\% & 4.6\% \\

CP
 & 34\% & 23.5\%
 & 20\% & 17.5\%
 & 32\% & 4.6\%
 & 47\% & 42.3\%
 & 30\% & \textbf{16.2\%} \\
 
TH
 & \textbf{40\%} & \textbf{36.9\%}
 & \textbf{26\%} & \textbf{24.8\%}
 & \textbf{37\%} & \textbf{6.5\%}
 & \textbf{58\%} & \textbf{49.6\%}
 & \textbf{32\%} & 15.4\% \\

\midrule
TH + CP + BG
 & \textbf{50\%} & \textbf{47.7\%}
 & \textbf{34\%} & \textbf{33.5\%}
 & 43\% & \textbf{8.5\%}
 & \textbf{62\%} & \textbf{60.0\%}
 & \textbf{42\%} & \textbf{21.2\%} \\

TH + CP + LT
 & 48\% & 44.2\%
 & 31\% & 27.9\%
 & \textbf{46\%} & 7.3\%
 & 60\% & 54.6\%
 & 39\% & 20.0\% \\

TH + CP + TD
 & 43\% & 40.0\%
 & 29\% & 26.3\%
 & 38\% & 6.5\%
 & 56\% & 52.7\%
 & 32\% & 16.5\% \\

\midrule
All Factors
 & \textbf{54\%} & \textbf{49.7\%}
 & \textbf{44\%} & \textbf{41.0\%}
 & \textbf{49\%} & \textbf{11.5\%}
 & \textbf{65\%} & \textbf{63.1\%}
 & \textbf{52\%} & \textbf{23.8\%} \\
\bottomrule
\end{tabular}
    }
\end{table}
\subsubsection{Effects of Domain Randomization}

To analyze the role of domain randomization, we decompose the five randomization factors into two categories: appearance factors, including Background (BG), Lighting (LT), and Table Distractor (TD); and spatial factors, including Table Height (TH) and Camera Position (CP). In this experiment, all randomization is applied in an episode-wise manner, where parameters are sampled once at scene initialization and kept fixed throughout each rollout.

~\cref{tab:factorized_results} reports the success rates under Simulation OOD and zero-shot real-world evaluation across different domain randomization factors. Training with clean data leads to poor generalization, resulting in low success rates in both Sim-OOD and real environments. Introducing individual randomization factors consistently improves performance across tasks.

\textbf{Lesson 1: Spatial randomization is the primary driver of Sim2Real generalization.}
Among all factors, spatial perturbations such as camera pose (CP) and table height (TH) provide the largest performance gains, significantly improving success rates across both Sim-OOD and real-world evaluations. In contrast, appearance-level changes such as background (BG) and lighting (LT) yield smaller improvements when applied individually. This suggests that variations affecting geometric relationships and viewpoint are more critical for robust policy learning than purely visual changes.

\textbf{Lesson 2: Appearance randomization provides complementary benefits when combined with spatial perturbations.}
Although background and lighting perturbations alone produce relatively modest gains, combining them with spatial randomization further improves performance. In particular, configurations such as TH + CP + BG and TH + CP + LT consistently outperform single-factor settings across most tasks. The best results are achieved when all factors are applied simultaneously, yielding the highest success rates in both Sim-OOD and real-world evaluations.

These results indicate that while spatial variations play a dominant role in improving policy robustness, appearance-level randomization helps cover additional visual discrepancies between simulation and reality when combined with spatial perturbations.

\begin{table}[t]
\caption{Results of randomization granularity. $\Delta$ denotes the gain of frame-wise over episode-wise.}
\label{tab:render_granularity}
\centering
\resizebox{\columnwidth}{!}{
\begin{tabular}{lcc|cc|cc|cc|cc}
\toprule
 & \multicolumn{2}{c|}{Click Bell}
 & \multicolumn{2}{c|}{Place Empty Cup}
 & \multicolumn{2}{c|}{Beat Block Hammer}
 & \multicolumn{2}{c|}{Stack Bowls Two}
 & \multicolumn{2}{c}{Pick Dual Bottles} \\
Data Setting
 & Sim-OOD & Real
 & Sim-OOD & Real
 & Sim-OOD & Real
 & Sim-OOD & Real
 & Sim-OOD & Real \\
\midrule

Episode-wise CP
 & 34\% & 23.5\%
 & 20\% & 17.5\%
 & 32\% & 4.6\%
 & 47\% & 42.3\%
 & 30\% & 16.2\% \\

Frame-wise CP
 & 37\% & 31.9\%
 & 27\% & 25.6\%
 & 34\% & 7.3\%
 & 51\% & 49.2\%
 & 34\% & 19.6\% \\

\rowcolor{red!15}
$\Delta$ (Frame $-$ Episode)
 & 3.0\% & 8.4\%
 & 7.0\% & 8.1\%
 & 2.0\% & 2.7\%
 & 4.0\% & 6.9\%
 & 4.0\% & 3.4\% \\

\midrule

Episode-wise BG
 & 23\% & 11.5\%
 & 15\% & 10.2\%
 & 30\% & 2.7\%
 & 42\% & 40.4\%
 & 27\% & 12.3\% \\

Frame-wise BG
 & 31\% & 15.0\%
 & 27\% & 23.1\%
 & 41\% & 7.7\%
 & 53\% & 45.0\%
 & 32\% & 14.2\% \\

\rowcolor{red!15}
$\Delta$ (Frame $-$ Episode)
 & 8.0\% & 3.5\%
 & 12.0\% & 12.9\%
 & 11.0\% & 5.0\%
 & 11.0\% & 4.6\%
 & 5.0\% & 1.9\% \\

\midrule

Episode-wise LT
 & 22\% & 12.3\%
 & 12\% & 8.6\%
 & 29\% & 4.6\%
 & 43\% & 31.5\%
 & 24\% & 7.3\% \\

Frame-wise LT
 & 23\% & 14.6\%
 & 13\% & 9.0\%
 & 30\% & 5.0\%
 & 43\% & 32.6\%
 & 26\% & 7.3\% \\

\rowcolor{red!15}
$\Delta$ (Frame $-$ Episode)
 & 1.0\% & 2.3\%
 & 1.0\% & 0.4\%
 & 1.0\% & 0.4\%
 & 0.0\% & 1.1\%
 & 2.0\% & 0.0\% \\

\bottomrule
\end{tabular}
}
\end{table}

\subsubsection{Effects of Randomization Granularity}

Building upon the episode-wise setting used in Experiment 1, we next investigate whether increasing the temporal frequency of randomization further improves zero-shot Sim2Real transfer. Specifically, we consider episode-wise and frame-wise strategies as two representative boundary cases of temporal stochasticity. In this study, we limit frame-wise randomization to background textures, lighting configuration, and camera position. Factors that directly affect task feasibility, including distractor placement and table height, remain episode-wise to preserve task stability, maintain consistent task planning, and avoid disrupting motion execution during rollout.

\textbf{Lesson 3: Granularity matters.}
Applying randomization at the frame level consistently outperforms episode-wise randomization across tasks (~\cref{tab:render_granularity}). 
Frame-wise perturbations introduce continuous variation within each rollout, preventing the policy from overfitting to static scene configurations. 
This effect is particularly evident for background (BG) and camera pose (CP), where frame-wise randomization improves real-world success rates by roughly $3$–$8\%$ for CP and up to around $13\%$ for BG, with similarly consistent gains observed under Sim-OOD evaluation. 
In contrast, lighting perturbations provide comparatively smaller improvements, typically around $0$–$2\%$, suggesting that illumination changes alone contribute less to generalization when applied at a finer temporal granularity.

\subsection{Rendering Fidelity for Sim2Real Generalization}
\begin{figure}[tbp]
\centering
\begin{subfigure}{0.49\linewidth}
    \centering
    \includegraphics[width=\linewidth]{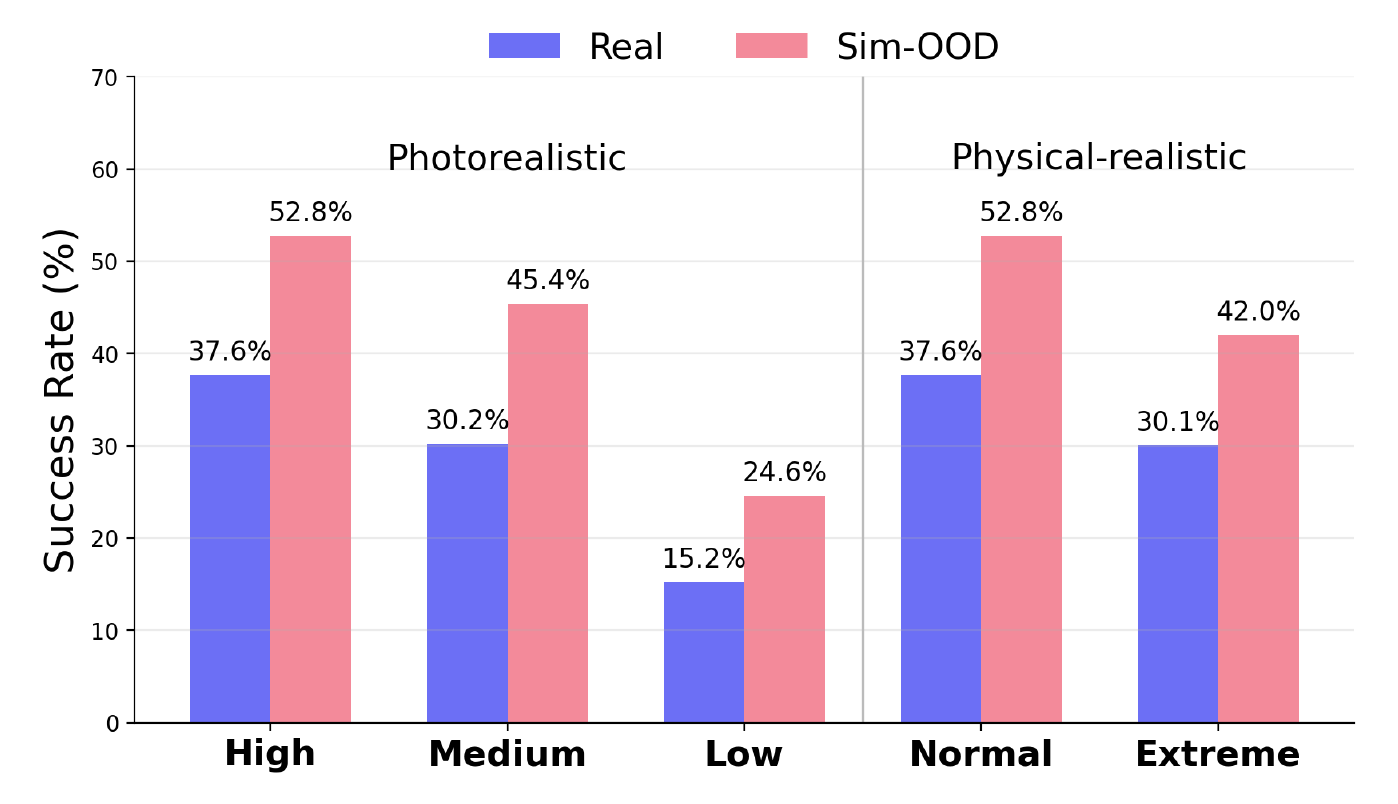}
    \caption{}
    \label{fig:left}
\end{subfigure}
\hfill
\begin{subfigure}{0.49\linewidth}
    \centering
    \includegraphics[width=\linewidth]{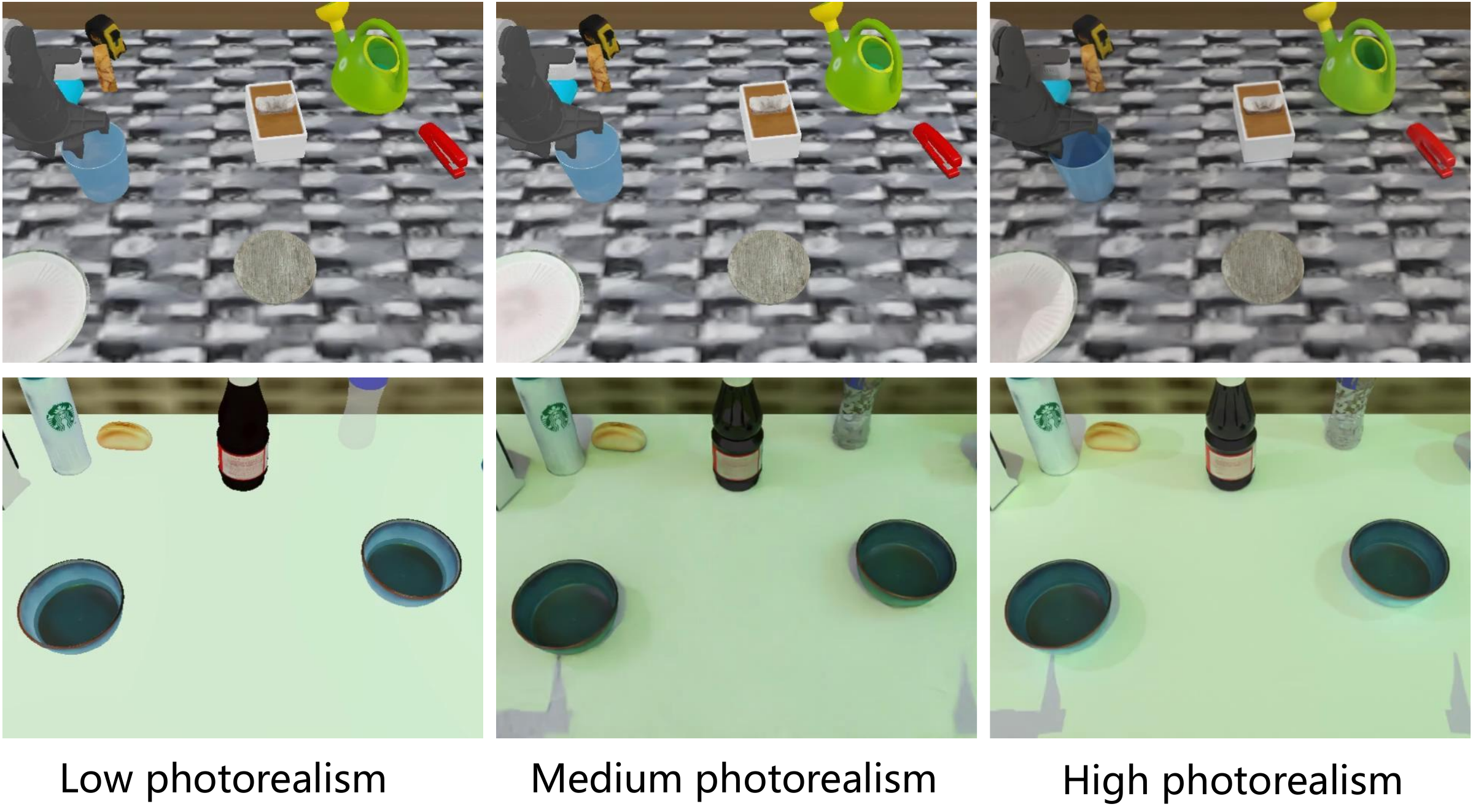}
    \caption{}
    \label{fig:right}
\end{subfigure}
\caption{Rendering fidelity analysis. (a) Quantitative results showing the impact of photorealism and physical realism on Sim-OOD and real-world success rates. (b) Example renderings under different photorealism levels (Low, Medium, High).}
\label{fig:two_figures}
\end{figure}

We next investigate whether increasing rendering fidelity improves zero-shot Sim2Real transfer (\cref{fig:right}). 
We compare three rendering presets that differ in ray-tracing activation and sampling configuration:
The \textbf{Low} setting disables RT and uses low sampling and path depth. 
The \textbf{Medium} setting enables RT with the same sampling and path depth as the Low setting, while introducing additional rendering effects. 
The \textbf{High} setting further increases the sampling rate and path depth to achieve higher visual fidelity.

In addition to visual fidelity, we also consider physical realism. 
To examine the effect of physics fidelity, we construct a setting with intentionally distorted dynamics by modifying key physical parameters, including gravitational acceleration, static and dynamic friction coefficients, and restitution. 
These parameters are set to relatively extreme values to reduce the physical realism of the simulator, allowing us to evaluate how violations of realistic dynamics affect VLA training and Sim2Real transfer.

\textbf{Lesson 4: Simulation fidelity enhances Sim2Real transfer.}
~\cref{fig:left} reports the performance comparison across five tasks under different rendering fidelity settings. 
Increasing photorealism consistently improves real-world success rates, indicating that more realistic visual appearance helps the policy learn representations that transfer better to the physical environment. 
However, the improvement becomes marginal beyond the medium setting, suggesting that moderate photorealism already captures most transferable visual cues.

The figure also shows that improving physical realism leads to certain gains in Sim2Real performance for VLA models. However, the improvement is smaller compared with that obtained from increasing photorealism, indicating that visual fidelity plays a more dominant role in facilitating Sim2Real transfer.
\subsection{Reinforcement Learning for Sim2Real Generalization}
Finally, we evaluate the effect of reinforcement learning (RL) fine-tuning on zero-shot Sim2Real transfer. Following the design in SimpleVLA~\cite{simplevla}, we modify the action prediction head of OpenVLA-OFT by replacing the original MLP-based continuous action regression with the LLaMA2 output head that generates discretized action tokens optimized with a cross-entropy loss.

Based on this modification, we first re-train an SFT policy using 100 demonstrations per task, which serves as the initialization for all RL experiments. Starting from this SFT checkpoint, we further perform RL fine-tuning in simulation under different rollout settings. We compare three training variants: (1) SFT trained in a clean simulation environment, (2) SFT followed by RL fine-tuning using rollouts collected in a clean simulation environment, and (3) SFT followed by RL fine-tuning using rollouts collected under domain randomization.

\textbf{Lesson 5: RL enhances robustness to distribution shifts.}
As shown in~\cref{tab:rl_results}, even when RL is trained under clean simulation conditions, it substantially improves performance over the SFT baseline, increasing the average real-world success rate from $5.6\%$ to $33.4\%$. 
Notably, this clean RL setting already approaches the performance achieved by SFT combined with domain randomization as shown in~\cref{tab:factorized_results}, suggesting that RL itself introduces stronger policy generalization beyond supervised imitation. 
Furthermore, when domain randomization is incorporated during RL rollouts, the performance further improves to $42.8\%$ in real-world evaluation and $70.8\%$ in Sim-OOD, indicating that RL can effectively exploit diverse training experiences to further enhance Sim2Real robustness.
\begin{table}[t]
\caption{Effect of reinforcement learning (RL) and domain randomization (DR) on zero-shot Sim2Real performance across five tasks.}
\label{tab:rl_results}
\centering
\resizebox{\columnwidth}{!}{
\begin{tabular}{lcc|cc|cc|cc|cc}
\toprule
 & \multicolumn{2}{c|}{Click Bell}
 & \multicolumn{2}{c|}{Place Empty Cup}
 & \multicolumn{2}{c|}{Beat Block Hammer}
 & \multicolumn{2}{c|}{Stack Bowls Two}
 & \multicolumn{2}{c}{Pick Dual Bottles} \\
Training Variant
 & Sim-OOD & Real
 & Sim-OOD & Real
 & Sim-OOD & Real
 & Sim-OOD & Real
 & Sim-OOD & Real \\
\midrule
SFT
 & 11\% & 0\%
 & 23\% & 6.8\%
 & 15\% & 0\%
 & 33\% & 21.2\%
 & 10\% & 0\% \\

SFT + RL
 & 42\% & 36.9\%
 & 65\% & 34.8\%
 & 52\% & 9.6\%
 & 60\% & 59.2\%
 & 48\% & 26.5\% \\

SFT + RL + DR
 & \textbf{60\%} & \textbf{50.8\%}
 & \textbf{87\%} & \textbf{51.4\%}
 & \textbf{72\%} & \textbf{16.2\%}
 & \textbf{68\%} & \textbf{64.6\%}
 & \textbf{67\%} & \textbf{30.8\%} \\
\bottomrule
\end{tabular}}
\end{table}

\section{Conclusion}
In this work, we systematically study how different Sim2Real techniques affect the generalization of Vision-Language-Action (VLA) models. Our experiments show that spatial domain randomization plays a dominant role in improving robustness, while appearance randomization provides complementary benefits. We further find that frame-wise randomization is more effective than episode-wise perturbations, higher simulation fidelity enhances Sim2Real transfer, and reinforcement learning fine-tuning significantly improves robustness to distribution shifts.

In addition, we develop an online benchmarking platform that provides standardized protocols, experimental settings, and real-world deployment interfaces for bimanual robots, enabling reproducible evaluation and continued benchmarking of Sim2Real performance for VLA models.

\section*{Acknowledgments}

This work is supported in part by 
Shenzhen Science and Technology Program under grant KJZD20240903104008012, Shenzhen Science and Technology Program under grant ZDCY20250901113000001, CUHK-CUHK(SZ)-GDSTC Joint Collaboration Fund No. 2025A0505000053, 
and GuangDong Key Laboratory of Big Data Computing (2021B1212040002).
\bibliographystyle{splncs04}
\bibliography{main}

\clearpage
\setcounter{page}{1}
\appendix
\onecolumn

\begin{center}

{\bf {\Large Appendix}} 
\end{center}
\thispagestyle{plain}

\section{Implementation Details}
\subsection{Model Architecture Details}
\begin{figure}[t]
    \centerline{\includegraphics[width=1\linewidth]{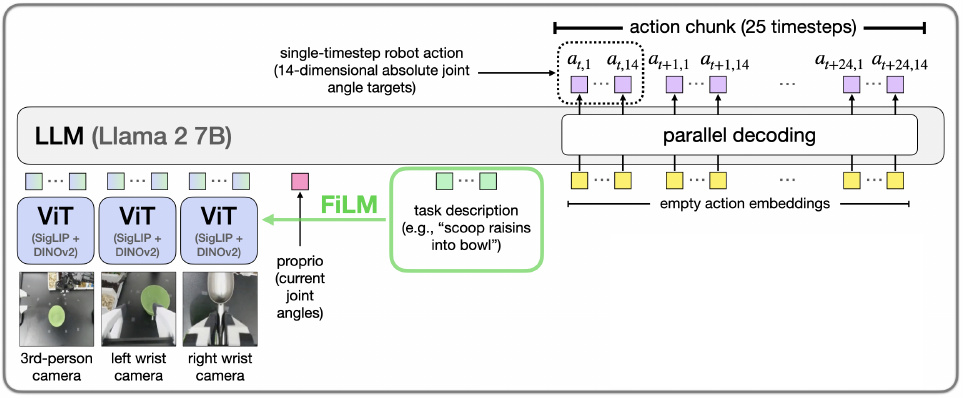}}
    \caption{\small Model architecture of OpenVLA-OFT.}
    \label{fig:arch}
\end{figure}

The model architecture is based on OpenVLA-OFT~\cite{openvla-oft}. In the original OpenVLA~\cite{openvla} framework, a single image and a language instruction are jointly processed to predict robot actions. Visual features are extracted by a fused vision backbone composed of SigLIP~\cite{zhai2023sigmoid} and DINOv2~\cite{oquab2023dinov2} vision transformers, projected into the language embedding space through an MLP projector, and concatenated with language embeddings before being processed by the Llama-2 decoder~\cite{Touvron2023Llama2O}. Building on OpenVLA, OpenVLA-OFT further projects robot proprioceptive state into the language embedding space using a 2-layer MLP with GELU activation, replaces causal attention with bidirectional attention for parallel decoding, substitutes the original language model output head with a 4-layer MLP with ReLU activation to directly predict continuous actions instead of discrete tokens, and predicts chunks of $K$ actions at each step rather than a single-timestep action. In addition, FiLM~\cite{perez2018film} modules are introduced to modulate visual features in both the SigLIP and DINOv2 vision transformers using the averaged task language embedding. The complete architecture is illustrated in~\cref{fig:arch}.
\subsubsection{Architecture Adaptation for RL finetuning.}
For the RL experiments, we follow SimpleVLA~\cite{simplevla} and retain only the parallel decoding and action chunking designs. Instead of replacing the language model head with an MLP for continuous action regression, the model keeps the original LLaMA2 output head to generate discrete action tokens and is trained with a cross-entropy loss. This modification enables efficient rollout by allowing the policy to sample actions as discrete tokens using standard language-model decoding procedures. The supervised fine-tuning stage uses the same hyperparameter settings as the official OpenVLA implementation.

\subsection{Training Details}
SFT and RL fine-tuning are conducted on 8$\times$NVIDIA H100 80GB GPUs. Specifically, the SFT stage employs LoRA with rank 32, a batch size of 64, and an initial learning rate of $5 \times 10^{-4}$, with learning rate decay starting after 15{,}000 gradient steps, and the total number of training steps varies from 20{,}000 to 30{,}000 depending on the task horizon. The RL fine-tuning stage is initialized from the SFT checkpoint, with a training batch size of 64, a validation batch size of 256, and 8 rollout samples collected per prompt. The actor learning rate is set to $5 \times 10^{-6}$ with constant warmup. The PPO mini-batch size is 128, the trajectory mini-batch size is 8, and gradient clipping is set to 1. The PPO clipping thresholds are $\epsilon_L = 0.2$ and $\epsilon_H = 0.28$, and the rollout temperature is set to $1.6$. We use an action token length of 14 and an action chunk length of 25. The maximum prompt length is 256 and the maximum response length is 128. RL training runs for 50 epochs, using the GRPO advantage estimator with zero KL regularization. To improve training efficiency, rollout samples are filtered by accuracy, retaining only those with success rates between 0.1 and 0.9.

\section{Simulation Datasets}
\subsection{Task Description}

We evaluate our method on five bimanual manipulation tasks from the RoboTwin 2.0 benchmark~\cite{Robotwin2.0}.~\cref{tab:tasks} summarizes the task descriptions and their average step counts.

\begin{table}[ht]
    \centering
    \caption{Task descriptions and average step counts for the five evaluation tasks.}
    \label{tab:tasks}
    \begin{tabular}{l l c}
        \toprule
        \textbf{Task Name} & \textbf{Description} & \textbf{Avg. Steps} \\
        \midrule
        Stack Bowls Two & Stack two bowls on top of each other & 313 \\
        Beat Block Hammer   & Use the hammer to strike the block. & 113 \\
        Pick Dual Bottles & Pick up one bottle with each arm. & 127 \\
        Click Bell & Click the bell's top center on the table. & 85 \\
        Place Empty Cup & Use an arm to place the empty cup on the coaster. & 174 \\
        \bottomrule
    \end{tabular}
\end{table}
\subsection{Domain randomization Details}
Domain randomization includes cluttered scenes, random lighting, table height variation, unseen language instructions, randomized background textures, and head camera pose perturbation.
\begin{itemize}
    \item \textbf{Scene Clutter}: Randomly add task-irrelevant distractors from RoboTwin-OD (731 objects, 147 categories) with collision-aware placement.
    
    \item \textbf{Background Textures}: Apply random textures from a curated library of 11,000 high-quality surface.
    
    \item \textbf{Lighting}: Randomize light color, type, intensity, and position within physically plausible bounds.
    
    \item \textbf{Tabletop Height}: Uniformly vary table height up to \(\pm 3\) cm.

    \item \textbf{Camera Pose}: Apply random displacement up to \(0.01\) m from nominal mounting position.
\end{itemize}

\subsection{Fedility Rendering Details}

\subsubsection{Photorealistic Rendering Details.}

We configure three levels of rendering fidelity to systematically vary visual realism and computational cost across experiments:

\begin{itemize}
    \item \textbf{High fidelity}: Ray-tracing renderer with 32 samples per pixel (spp), ray-tracing path depth of 8, and OIDN denoising enabled.
    
    \item \textbf{Medium fidelity}: Ray-tracing renderer with 4 spp, path depth of 4, and denoiser disabled.
    
    \item \textbf{Low fidelity}: Standard (non-ray-tracing) renderer with 4 spp and path depth of 4, and denoiser disabled.
\end{itemize}
\subsubsection{Physical-Realistic Rendering Details.}
We construct an intentionally extreme simulation environment to evaluate the sensitivity of VLA policies to physical dynamics. Specifically, friction coefficients are reduced to $0.001$ (from $0.5$), restitution is increased to $0.2$ (from $0.0$), and gravity is lowered to $0.05\,\mathrm{m/s^2}$ (from $9.81\,\mathrm{m/s^2}$). These values create a substantial deviation from real-world physics, producing object behaviors that differ noticeably from those encountered in realistic environments. Rather than modeling plausible physical conditions, this setup serves as a controlled stress test that amplifies dynamic discrepancies, allowing us to systematically examine how sensitive VLA policies are to changes in environmental dynamics and to better characterize their robustness under severe Sim-to-Real mismatch.
\begin{table}[t]
\caption{Factorized domain randomization results under zero-shot real-world evaluation on five tasks.}
\label{tab:two_tasks_real}
\centering
\scriptsize
\setlength{\tabcolsep}{3.5pt}
\resizebox{\columnwidth}{!}{
\begin{tabular}{llccccc}
\toprule
\multirow{2}{*}{Task} & \multirow{2}{*}{Data setting} & \multicolumn{5}{c}{Real setting} \\
\cmidrule(lr){3-7}
& & Base & BG & Light & Distractor & Object \\
\midrule
\multirow{10}{*}{Click Bell}
& Clean          & 1/20  & 1/60  & 3/60  & 0/60  & 2/60  \\
& BG             & 6/20  & 8/60  & 6/60  & 4/60   & 6/60   \\
& LT             & 6/20  & 5/60  & 10/60  & 4/60   & 7/60  \\
& TD             & 2/20  & 1/60  & 1/60  & 3/60  & 1/60  \\
& CP             & 10/20  & 12/60  &  13/60 & 9/60  & 17/60 \\
& TH             & 14/20  & 20/60  & 22/60 & 16/60  & 24/60  \\
& TH + CP + BG   & 7/20  & 13/60 & 14/60 & 10/60 & 11/60 \\
& TH + CP + LT   & 7/20  & 11/60 & 14/60 & 9/60  & 11/60 \\
& TH + CP + TD   & 6/20  & 8/60  & 9/60  &10/60 & 10/60 \\
& All Factors    & \textbf{15/20} & \textbf{30/60} & \textbf{39/60} & \textbf{14/60} & \textbf{31/60} \\
\midrule
\multirow{10}{*}{Place Empty Cup}
& Clean         & 3/40  & 9/120  & 7/120  & 2/120  & 7/120 \\
& BG            & 8/40  & 16/120 & 16/120 & 3/120  & 10/120 \\
& LT            & 6/40  & 9/120  & 17/120 & 3/120  & 10/120 \\
& TD            & 4/40  & 8/120  & 12/120 & 7/120  & 5/120 \\
& CP            & 13/40 & 12/120 & 35/120 & 12/120 & 22/120 \\
& TH            & 16/40 & 24/120 & 43/120 & 14/120 & 32/120 \\
& TH + CP + BG  & \textbf{23/40} & \textbf{42/120} & 51/120 & 23/120 & 35/120 \\
& TH + CP + LT  & 19/40 & 29/120 & 50/120 & 12/120 & 35/120 \\
& TH + CP + TD  & 21/40 & 28/120 & 32/120 & 26/120 & 30/120 \\
& All Factors   & 22/40 & \textbf{42/120} & \textbf{53/120} & \textbf{38/120} & \textbf{58/120} \\
\midrule
\multirow{10}{*}{Beat Block Hammer}
& Clean          & 0/20  & 0/60  & 0/60  & 0/60 & 0/60 \\
& BG             & 1/20  & 3/60  & 2/60  & 0/60 & 1/60 \\
& LT             & 2/20  & 3/60  & 6/60  & 0/60 & 0/60 \\
& TD             & 0/20  & 0/60  & 0/60  & 0/60 & 0/60 \\
& CP             & 2/20  & 4/60  & 4/60  & 0/60 & 2/60 \\
& TH             & 3/20  & 4/60  & 6/60  & 0/60 & 4/60 \\
& TH + CP + BG   & 4/20  & 6/60  & 7/60  & 0/60 & 5/60 \\
& TH + CP + LT   & 3/20  & 9/60  & 9/60  & 0/60 & 3/60 \\
& TH + CP + TD   & 3/20  & 4/60  & 5/60  & 0/60 & 5/60 \\
& All Factors    & \textbf{3/20} & \textbf{9/60} & \textbf{10/60} & \textbf{0/60} & \textbf{8/60} \\
\midrule
\multirow{10}{*}{Stack Bowls Two}
& Clean          & 9/20  & 9/60  & 19/60 & 12/60 & 19/60 \\
& BG             & 9/20  & 24/60 & 29/60 & 16/60 & 27/60 \\
& LT               & 9/20  & 18/60 & 23/60 & 13/60 & 19/60 \\
& TD               & 8/20  & 12/60 & 17/60 & 18/60 & 16/60 \\
& CP               & 11/20 & 23/60 & 31/60 & 17/60 & 28/60 \\
& TH               & 12/20 & 30/60 & 33/60 & 19/60 & 35/60 \\
& TH + CP + BG     & 13/20 & 38/60 & 40/60 & 28/60 & 37/60 \\
& TH + CP + LT     & 13/20 & 32/60 & 35/60 & 25/60 & 37/60 \\
& TH + CP + TD     & 12/20 & 33/60 & 32/60 & 26/60 & 34/60 \\
& All Factors   & \textbf{14/20} & \textbf{40/60} & \textbf{42/60} & \textbf{29/60} & \textbf{39/60}\\
\midrule
\multirow{10}{*}{Pick Dual Bottles}
& Clean          & 1/20  & 1/60  & 1/60  & 0/60  & 1/60  \\
& BG             & 4/20  & 10/60 & 7/60  & 5/60  & 6/60  \\
& LT             & 3/20  & 3/60  & 6/60  & 3/60  & 4/60  \\
& TD             & 2/20  & 2/60  & 2/60  & 4/60  & 2/60  \\
& CP             & 5/20  & 9/60  & 10/60 & 8/60  & 10/60 \\
& TH             & 5/20  & 9/60  & 11/60 & 6/60  & 9/60  \\
& TH + CP + BG   & \textbf{7/20}  & 13/60 & 14/60 & \textbf{10/60} & 11/60 \\
& TH + CP + LT   & \textbf{7/20}  & 11/60 & 14/60 & 9/60  & 11/60 \\
& TH + CP + TD   & 6/20  & 8/60  & 9/60  & \textbf{10/60} & 10/60 \\
& All Factors    & \textbf{7/20} & \textbf{15/60} & \textbf{17/60} & \textbf{10/60} & \textbf{13/60} \\
\bottomrule
\end{tabular}
}

\end{table}

\begin{table}[t]
\caption{Frame-wise domain randomization results under zero-shot real-world evaluation across five tasks.}
\label{tab:freq_real}
\centering
\scriptsize
\setlength{\tabcolsep}{3.5pt}
\resizebox{\columnwidth}{!}{
\begin{tabular}{llccccc}
\toprule
\multirow{2}{*}{Task} & \multirow{2}{*}{Data setting} & \multicolumn{5}{c}{Real setting} \\
\cmidrule(lr){3-7}
& & Base & BG & Light & Distractor & Object \\
\midrule
\multirow{3}{*}{Click Bell}
&Frame-wise CP & 10/20 & 16/60 & 22/60 & 12/60 & 23/60 \\
&Frame-wise BG  & 4/20  & 11/60 & 8/60  & 6/60  & 10/60 \\
&Frame-wise LT & 4/20  & 6/60  & 12/60 & 5/60  & 11/60 \\
\midrule
\multirow{3}{*}{Place Empty Cup}
&Frame-wise CP & 18/40 & 24/120 & 39/120 & 19/120 & 33/120 \\
&Frame-wise BG  & 10/40 & 29/120 & 30/120 & 23/120 & 28/120 \\
&Frame-wise LT  &  6/40 &  9/120 & 19/120 &  3/120 & 10/120 \\
\midrule
\multirow{3}{*}{Beat Block Hammer}
& Frame-wise CP   & 3/20 & 5/60 & 6/60 & 0/60 & 5/60 \\
& Frame-wise BG   & 3/20 & 6/60 & 5/60 & 0/60 & 6/60  \\
& Frame-wise LT   & 2/20 & 4/60 & 5/60 & 0/60 & 2/60  \\
\midrule
\multirow{3}{*}{Stack Bowls Two}
& Frame-wise CP   & 15/20 & 25/60 & 34/60 & 22/60 & 32/60 \\
& Frame-wise BG   & 10/20 & 30/60 & 31/60 & 17/60 & 29/60\\
& Frame-wise LT   & 9/20  & 18/60 & 26/60 & 13/60 & 19/60 \\
\midrule
\multirow{3}{*}{Pick Dual Bottles}
& Frame-wise CP   & 6/20  & 13/60 & 14/60 & 7/60  & 11/60 \\
& Frame-wise BG   & 4/20  & 12/60 & 7/60  & 5/60  & 9/60  \\
& Frame-wise LT   & 3/20  & 2/60  & 8/60  & 2/60  & 4/60  \\
\bottomrule
\end{tabular}
}
\end{table}

\begin{table}[t]
\caption{Impact of photorealistic and physical-realistic fidelity on zero-shot real-world policy generalization. Default denotes high-fidelity rendering with ray-tracing enabled and standard physics settings. All experiments are trained using the All Factors domain randomization configuration.}
\label{tab:rendering}
\centering
\scriptsize
\setlength{\tabcolsep}{3.5pt}
\resizebox{\columnwidth}{!}{
\begin{tabular}{llccccc}
\toprule
\multirow{2}{*}{Task} & \multirow{2}{*}{Data setting} & \multicolumn{5}{c}{Real setting} \\
\cmidrule(lr){3-7}
& & Base & BG & Light & Distractor & Object \\
\midrule
\multirow{3}{*}{Click Bell}
& Default   & \textbf{15/20} & \textbf{30/60} & \textbf{39/60} & \textbf{14/60} & \textbf{31/60} \\
& Medium Visual & 14/20 & \textbf{30/60} & 32/60 & \textbf{14/60} & 30/60 \\
& Low Visual    & 3/20 & 7/60 & 9/60 & 5/60 & 10/60 \\
& Extreme Physics & 12/20 & 24/60 & 32/60 & 11/60 & 26/60 \\
\midrule
\multirow{3}{*}{Place Empty Cup}
& Default  & \textbf{22/40} & \textbf{42/120} & \textbf{53/120}& \textbf{38/120} & \textbf{58/120} \\
& Medium Visual & 12/40 & 23/120 & 30/120 & 20/120 & 35/120 \\
& Low Visual &  8/40 & 15/120 & 19/120 & 13/120 & 21/120 \\
&Extreme Physics    & 20/40 & 37/120 & 46/120 & 34/120 & 51/120 \\
\midrule
\multirow{3}{*}{Beat Block Hammer}
& Default  & \textbf{3/20} & \textbf{9/60} & \textbf{10/60}& 0/60 & \textbf{8/60} \\
& Medium Visual & \textbf{3/20} & 7/60 & 5/60 & 0/60 & 5/60 \\
& Low Visual & 0/20 & 0/60 & 0/60 & 0/60 & 0/60 \\
& Extreme Physics & 1/20 & 2/60 & 2/60 & 0/60 & 2/60 \\
\midrule
\multirow{3}{*}{Stack Bowls Two}
& Default  & \textbf{14/20} & \textbf{40/60} & \textbf{42/60}& \textbf{29/60} & \textbf{39/60} \\
& Medium Visual & 12/20 & 32/60 & 34/60 & 28/60 & 37/60 \\
& Low Visual & 10/20 & 24/60 & 27/60 & 21/60 & 27/60 \\
& Extreme Physics & \textbf{14/20} & 37/60 & 41/60 & \textbf{29/60} & 33/60 \\
\midrule
\multirow{3}{*}{Pick Dual Bottles}
& Default  & \textbf{7/20} & \textbf{15/60} & \textbf{17/60}& \textbf{10/60} & \textbf{13/60} \\
& Medium Visual & 5/20 & 12/60 & 13/60 & 8/60 & 12/60 \\
& Low Visual & 2/20 & 4/60 & 4/60 & 2/60 & 4/60 \\
& Extreme Physics & 3/20 & 7/60 & 8/60 & 5/60 & 8/60 \\
\bottomrule
\end{tabular}
}
\end{table}

\begin{table}[t]
\caption{RL fine-tuning results under zero-shot real-world evaluation across five tasks.}
\label{tab:RL}
\centering
\scriptsize
\setlength{\tabcolsep}{3.5pt}
\resizebox{\columnwidth}{!}{
\begin{tabular}{llccccc}
\toprule
\multirow{2}{*}{Task} & \multirow{2}{*}{Data setting} & \multicolumn{5}{c}{Real setting} \\
\cmidrule(lr){3-7}
& & Base & BG & Light & Distractor & Object \\
\midrule
\multirow{3}{*}{Click Bell}
& SFT   & 0/20 & 0/60 & 0/60 &0/60 & 0/60 \\
& SFT + RL & 11/20 & 26/60 & 28/60 & 12/60 & 19/60 \\
& SFT + RL + DR    & \textbf{13/20} & \textbf{32/60} & \textbf{35/60} & \textbf{23/60} & \textbf{29/60} \\
\midrule
\multirow{3}{*}{Place Empty Cup}
& SFT               & 6/40 & 7/120 & 12/120 & 3/120 & 7/120 \\
& SFT + RL          & 19/40 & 32/120 & 52/120 & 24/120 & 54/120 \\
& SFT + RL + DR     & \textbf{26/40} & \textbf{59/120} & \textbf{75/120} & \textbf{43/120} & \textbf{64/120} \\
\midrule
\multirow{3}{*}{Beat Block Hammer}
& SFT   & 0/20 & 0/60 & 0/60 &0/60 & 0/60 \\
& SFT + RL          & 4/20 & 8/60 & 8/60 & 0/60 & 5/60 \\
& SFT + RL + DR    & \textbf{5/20} & \textbf{12/60} & \textbf{14/60} & \textbf{1/60} & \textbf{10/60} \\
\midrule
\multirow{3}{*}{Stack Bowls Two}
& SFT   & 6/20 & 13/60 & 14/60 & 9/60 & 13/60 \\
& SFT + RL          & 14/20 & 35/60 & 41/60 & 28/60 & 36/60 \\
& SFT + RL + DR    & \textbf{16/20} & \textbf{38/60} & \textbf{45/60} & \textbf{31/60} & \textbf{38/60} \\
\midrule
\multirow{3}{*}{Pick Dual Bottles}
& SFT   & 0/20 & 0/60 & 0/60 & 0/60 & 0/60 \\
& SFT + RL          & 7/20 & 16/60 & 20/60 & 12/60 & 14/60 \\
& SFT + RL + DR    & \textbf{8/20} & \textbf{20/60} & \textbf{21/60} & \textbf{16/60} & \textbf{15/60} \\
\bottomrule
\end{tabular}
}
\end{table}
\section{Further Results}
\subsection{Real-World Evaluation Results}

All real-world experiments are conducted under a canonical base configuration to provide a controlled evaluation setup. In this base setting, the scene uses a white background, no lighting variation, the same object instances as in training, and no table distractors. This configuration serves as the reference environment for evaluating the baseline performance of the policy.

\begin{figure}[t]
    \centerline{\includegraphics[width=1\linewidth]{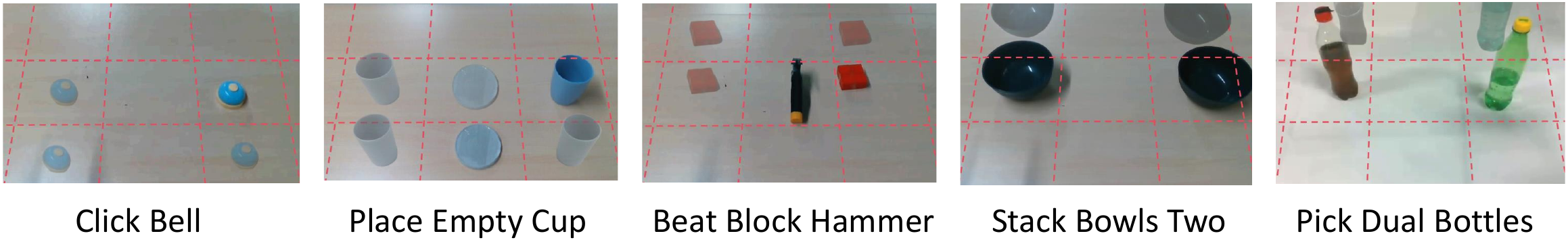}}
    \caption{Pose design for real-world evaluation across five manipulation tasks. The \textit{Place Empty Cup} task includes eight pose variations, while the other tasks use four poses variations.}
    \label{fig:grid}
\end{figure}

To evaluate the robustness of the policy under different environmental conditions, we introduce several variations including background changes, lighting variation, object replacement, and the presence of table distractors. Each variation is evaluated independently to analyze how different factors affect real-world performance. For every evaluation condition, the object placement is systematically varied using a predefined $3\times3$ grid layout(see~\cref{fig:grid}), which changes the spatial position of the object relative to the robot. 

~\cref{tab:two_tasks_real,tab:freq_real,tab:rendering,tab:RL} report detailed real-world evaluation results across different real-world environments, analyzing how policy performance changes with variations in visual appearance, physical fidelity, and RL fine-tuning. Among the evaluated factors, distractors introduce the largest performance degradation, while lighting variation has the smallest effect on success rate. 

\section{Sim-to-Real Benchmark Platform}
\begin{figure}[t]
    \centerline{\includegraphics[width=1\linewidth]{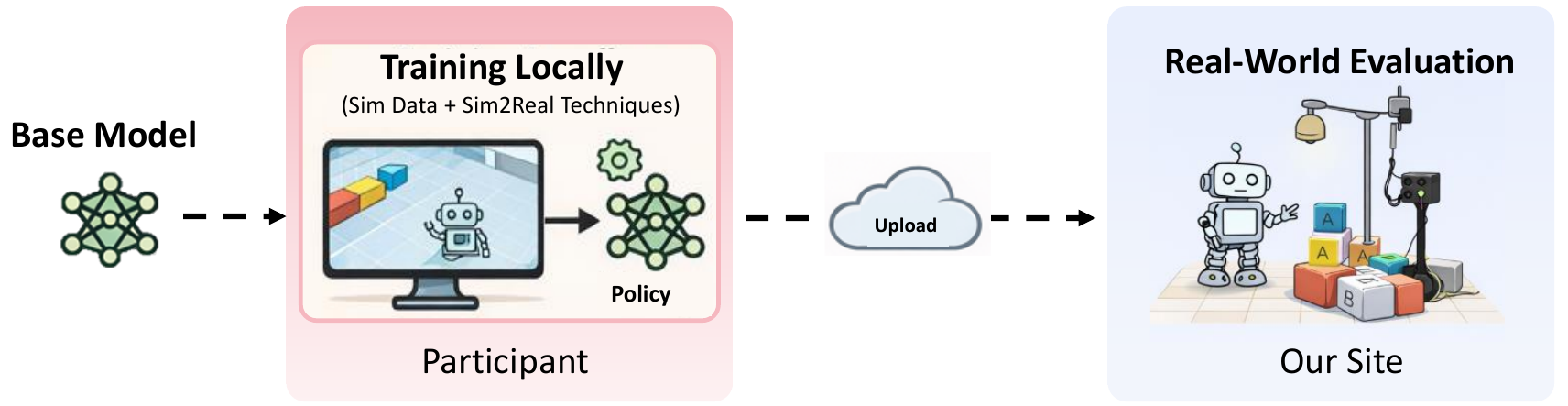}}
    \caption{Overview of the Sim-to-Real benchmark platform. Participants train policies locally using simulation data and Sim2Real techniques, upload the trained models, and the policies are evaluated under standardized real-world robotic setups at our site.}
    \label{fig:challenge}
\end{figure}

Inspired by the RoboChallenge~\cite{RoboChallenge} real-robot evaluation framework, we design a platform that enables researchers to evaluate policies trained in simulation on real robotic systems. As illustrated in~\cref{fig:challenge}, participants start from a base model and perform local training using simulated environments while applying Sim2Real techniques to improve transferability. The resulting policy is then uploaded to our centralized evaluation site, where it is executed on real robots under standardized experimental setups. This design allows researchers to develop and iterate policies locally without requiring direct access to hardware, while ensuring consistent and reproducible real-world evaluation across different methods.

\end{document}